\documentclass[journal]{IEEEtran}
\ifCLASSINFOpdf
\else
\fi
\usepackage{amsmath, amsthm, amssymb}
\usepackage{algorithm}
\usepackage{multirow}
\usepackage{array}
\usepackage{booktabs}
\usepackage{bm}
\usepackage{graphicx}
\usepackage{xcolor}
\usepackage{soul}
\usepackage{hyperref}
\usepackage{comment}
\usepackage{fixmath}

\usepackage{makecell}

\begin{document}
%
\title{ADA-Net: Attention-Guided Domain Adaptation Network with Contrastive Learning for Standing Dead Tree Segmentation Using Aerial Imagery}

\author{Mete Ahishali, Anis Ur Rahman, Einari Heinaro, and Samuli Junttila
\thanks{Mete Ahishali, Anis Ur Rahman, Einari Heinaro, and Samuli Junttila are with the School of Forest Sciences, Faculty of Science, Forestry and Technology, University of Eastern Finland, Finland (email: \textit{name.surname@uef.fi}).}}
\maketitle

\begin{abstract}
Information on standing dead trees is important for understanding forest ecosystem functioning and resilience but has been lacking over large geographic regions. Climate change has caused large-scale tree mortality events that can remain undetected due to limited data. In this study, we propose a novel method for segmenting standing dead trees using aerial multispectral orthoimages. Because access to annotated datasets has been a significant problem in forest remote sensing due to the need for forest expertise, we introduce a method for domain transfer by leveraging domain adaptation to learn a transformation from a \textit{source domain X} to \textit{target domain Y}. In this Image-to-Image translation task, we aim to utilize available annotations in the target domain by pre-training a segmentation network. When images from a new study site without annotations are introduced (source domain X), these images are transformed into the target domain. Then, transfer learning is applied by inferring the pre-trained network on domain-adapted images. In addition to investigating the feasibility of current domain adaptation approaches for this objective, we propose a novel approach called the Attention-guided Domain Adaptation Network (ADA-Net) with enhanced contrastive learning. Accordingly, the ADA-Net approach provides new \textit{state-of-the-art} domain adaptation performance levels outperforming existing approaches. We have evaluated the proposed approach using two datasets from Finland and the US. The USA images are converted to the Finland domain, and we show that the synthetic \textit{USA2Finland} dataset exhibits similar characteristics to the Finland domain images. This transformation has improved source domain standing dead tree segmentation by nearly $20\%$ and obtained a dice score of around $44\%$. The gap between the proposed and best-performing competing domain adaptation approach is $4.5\%$ in dice score for segmentation. Overall, this study is a pioneer work in adapting images for cross-site tree mortality mapping where both sites significantly differ in terms of terrain types, tree species, and geographical conditions, while the annotated data are available from one site only. The software implementation is shared at \href{https://github.com/meteahishali/ADA-Net}{https://github.com/meteahishali/ADA-Net} and the data is publicly available \footnote{The USA dataset along with the annotations is shared at \href{https://www.kaggle.com/datasets/meteahishali/aerial-imagery-for-standing-dead-tree-segmentation}{https://www.kaggle.com/datasets/meteahishali/aerial-imagery-for-standing-dead-tree-segmentation}.}.
\end{abstract}

\begin{IEEEkeywords}
Aerial imagery, contrastive learning, domain adaptation, generative adversarial networks, self-attention mechanism, standing dead tree segmentation.
\end{IEEEkeywords}

%
\IEEEpeerreviewmaketitle

\section{Introduction}
\IEEEPARstart{S}{egmentation} of standing dead trees plays a vital role in forest health monitoring, planning reforestation actions, and controlling the spread of wildfires. Forests are crucial for the natural environment by storing the majority of terrestrial carbon deposits, serving as habitats for many wildlife species, and regulating climates \cite{deng2024individual}. Many studies have been proposed to map trees using remote sensing \cite{deng2024individual, lin2015use, mcroberts2007remote} and, in particular, monitor standing dead trees \cite{putman2018automated, chiang2020deep, bright2014landsat} to understand forest ecosystems and their climate resilience further. In many forest monitoring tasks \cite{deng2024individual,mcroberts2007remote}, decision-making leveraging Light Detection and Ranging (LiDAR) data is preferable due to the rich structural information it delivers about individual trees, especially for dense canopies \cite{xiang2013automatic}. On the other hand, there are several drawbacks to using LiDAR sensors, including the cost, availability, and coverage. Compared to LiDAR, optical aerial imagery is a cost-efficient alternative for forest decision-making systems since aerial images are widely available and can be obtained using less complex equipment \cite{zhou2024automatic}. This accessibility provides a temporal dimension as aerial imagery is generally collected for extended periods, allowing for the inspection of temporal changes. Moreover, multispectral imagery offers spectral-rich information about the observed scene yielding improved multimodal analysis. Therefore, leveraging aerial images has been preferred by many studies including \cite{lin2015use,chiang2020deep}.

Recently, methods utilizing Convolutional Neural Networks (CNNs) have achieved new state-of-the-art performance levels in many computer vision applications such as object detection \cite{zou2023object}, classification \cite{krizhevsky2017imagenet, osen}, image enhancement \cite{r2c_gan}, and segmentation \cite{minaee2021image}. This trend has also impacted recent studies in remote sensing applications, and methods based on CNNs have been proposed for land use/land cover classification \cite{ahishali2021classification} and wildfire detection \cite{devecioglu2023improved}. Specifically, in the scope of forest health applications using remote sensing, there has been a growing interest in deep learning-based approaches utilizing aerial imagery. For instance, the study in \cite{chiang2020deep}, focusing on forest health monitoring, proposed a segmentation model using Mask Region-based CNN (Mask RCNN) in order to classify healthy and dead trees over Scotland using aerial images. Another study \cite{zhou2024automatic} detected standing dead trees using a You Only Look Once (YOLO) based object detector with optical remote sensing data. Accordingly, they aimed to specifically detect small-sized dead trees across elevated surfaces with mountainous terrains, where the detection performance drops significantly with traditional automated detection frameworks. In the study of \cite{adedapo2024early}, different object detectors were utilized for detecting brown spot needle blight disease in pine species over the United States by optical images captured via Unmanned Aerial Vehicles (UAV). A multi-class detection was performed to discriminate healthy, diseased, and already dead trees, leveraging RetinaNet and Single Shot Detector (SSD) models.

While satisfactory performance levels have been obtained using CNN-based deep learning approaches in previously mentioned applications, they have several disadvantages. The receptive field size of traditional convolutional layers is limited, and the existing relations/correlations within the image are not fully captured during the training process. Transformer models equipped with the so-called attention mechanism were proposed in \cite{vaswani2017attention} for the Natural Language Processing task. Models with attention layers can better learn relevant parts in given input data, yielding improved accuracies for the applications where the data have complex patterns. Later, the Vision Transformer (ViT) approach was proposed by \cite{dosovitskiy2021an}, extending the application of the attention mechanism for computer vision tasks. In ViT, the images are divided into smaller patches, and the importance of individual patches is exploited, yielding better accuracies in different classification \cite{chen2021crossvit, chen2021visformer} and regression \cite{fan2022sunet, xu2023cur} tasks. Several studies \cite{bazi2021vision, peng2020optical} have explored complex model structures incorporating attention layers in different remote sensing applications. The attention mechanism is substantially important in remote sensing since the data are usually acquired at high resolution; hence, there is a need for larger receptive fields to perform decision-making based on the most descriptive regions.

While the aforementioned CNN-based deep learning approaches have achieved satisfactory performance levels, they have one common major drawback: they require vast amounts of annotated data generally on a ``big data" scale for training, which is especially scarce for forest remote sensing applications. Moreover, transformer models tend to be more ``data-hungry" to provide superior performance levels compared to CNNs as discussed in \cite{bai2021transformers}. This is because their attention mechanisms consist of a significantly larger number of parameters. Employing an automated standing dead tree detection framework leveraging attention layers can be unfeasible for most cases because training data is scarce for remote sensing applications. The labeling process of such large amounts of data is not only resource-intensive, but it also requires specialized domain knowledge, causing a substantial burden on available human resources. Labeling for some tasks may even be impossible without field reference, such as tree species classification.

In this study, we propose segmenting standing dead trees by deploying domain adaptation and removing the need for data annotation in the source domain images. Domain adaptation focuses on learning a transformation between source and target domains. Usually, the target domain depicts the desired domain with well-annotated data for a specific task, whereas the source domain annotated samples are scarce. Therefore, performing a decision-making task directly over the source domain is challenging. The transformation from source to target domain can be learned using Image-to-Image translation approaches based on Generative Adversarial Networks (GANs) \cite{goodfellow2020generative}. As pioneering generative models, GANs have started to be widely used in many applications requiring generative modeling, such as data augmentation \cite{yamaguchi2020effective}, text-to-image synthesis \cite{kang2023scaling}, and style transfer \cite{jabbar2021survey, cut, cycle_gan}. Other different variants operating in supervised manner are proposed such as conditional Generative Adversarial Network (cGAN) models \cite{isola2017image} requiring paired samples. Subsequently, Cycle-Consistent Generative Adversarial Networks (Cycle-GANs) \cite{cycle_gan} are introduced where both forward and inverse transformations are learned between the source and target domains. For a successful Image-to-Image translation, it is important to identify common features in the domains while distinguishing the dissimilarities that have been partially achieved by GANs and Cycle-GANs. To enhance this capability, Contrastive Unpaired Translation (CUT) \cite{cut} has been proposed to introduce contrastive loss. The proposed solution compares individual sampled pixels from generator feature maps when fed with the domain-adapted and target domain images.

In this study, we propose a novel approach called the Attention-guided Domain Adaptation Network (ADA-Net). The proposed approach is an improved version of CUT-based models with attentional layers and enhanced contrastive learning. First, it is an unsupervised and blind domain transformation method that operates with unpaired training data. In other words, training samples from the source and target domains do not need to be aligned. The domain adaptation performance has been evaluated in a zero-shot segmentation experimental setup, where four band multispectral aerial images from the USA are transformed to mimic the images obtained in Finland. Accordingly, a deep segmentation network with self-attention layers, Flair U-Net \cite{flair_unet}, is trained using annotated data over the Finland site. When the USA images need to be segmented, these images are converted to create a \textit{USA2Finland} synthetic dataset. Next, the \textit{USA2Finland} dataset is segmented using the pre-trained Flair U-Net without further fine-tuning. Such a proposed zero-shot segmentation over the USA is especially beneficial for deep learning models with attention layers. Since no data annotation is needed in the source domain, it is possible to provide scalable solutions in standing dead tree segmentation without additional resource allocation in data annotation and engagement of domain-specific expertise. Overall, we summarize the significance of this work and our contributions as follows,
\begin{itemize}
    \item First, we investigate the feasibility of utilizing different Image-to-Image translation techniques for domain adaptation in cross-site standing dead tree segmentation applications. This approach will eliminate the need for source domain segmentation data annotation.
    
    \item A novel domain adaptation approach called ADA-Net is proposed. Thanks to the introduced patch-wise contrastive learning in the frequency domain representation, the generator model is expected to learn to associate similar patches together while discriminating them from dissimilar patches usually located further apart.
    
    \item Attention mechanism in the generator network of the ADA-Net approach enables the exploration of the aligned relationships within the scene. Therefore, the generated images are more contextually relevant to the source domain samples.
    
    \item The proposed approach has obtained new state-of-the-art domain adaptation performances improving the source domain standing dead tree segmentation by approximately $20\%$ and attaining $44\%$ in dice score. We show that the tree characteristics specifically indicate that alive and dead trees have been better preserved after the domain transformation than the previous methods.

    \item Compact network configurations have been followed in the ADA-Net model, contributing computational efficiency with fewer trainable parameters than existing methods.

    \item Finally, the annotated USA multispectral dataset is shared publicly with the research community. The dataset has been meticulously labeled by our group of forest health experts and represents a significant contribution as it will attract more research in this field.
\end{itemize}

The remainder of this paper is organized as follows. A detailed presentation of the proposed ADA-Net approach and the concept of using different domain adaptation approaches for zero-shot segmentation is provided in Section \ref{sec:methodology}. Next, an extensive set of experimental evaluations is presented in Section \ref{sec:experimental_results} with discussions on the limitations and computational complexity. Finally, Section \ref{sec:conclusion} concludes this study.

\section{Methodology}
\label{sec:methodology}

This section first presents the proposed ADA-Net approach for domain adaptation tasks. Next, we will detail other methods utilized in this study, which consist of CUT, FastCUT, and Cycle-GAN models.

\subsection{Proposed ADA-Net: Attention-Guided Domain Adaptation Network}
\label{sec:ada-net}

\begin{figure*}[t!]
    \centering
    \includegraphics[width=.97\linewidth]{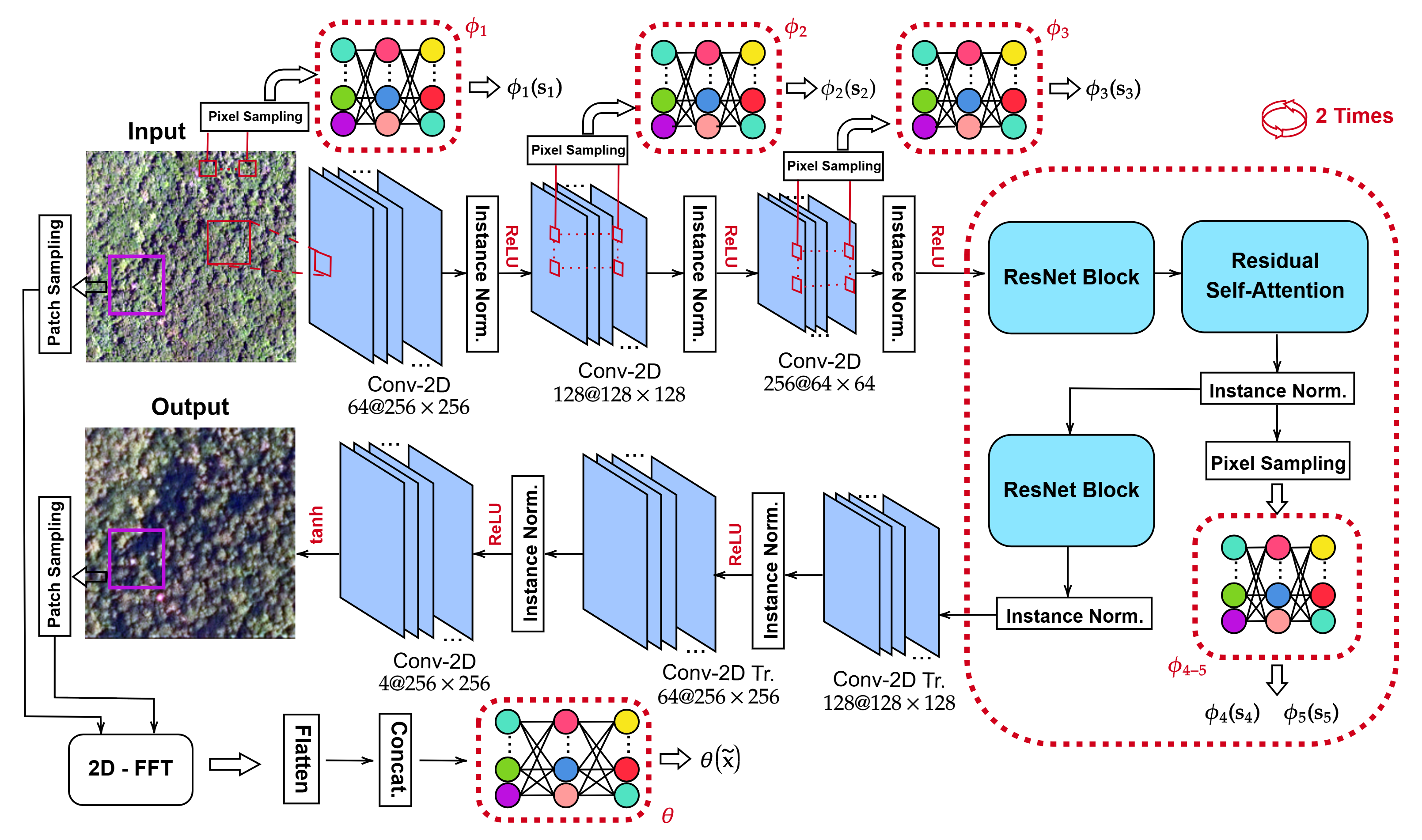}
    \caption{The generator network $G$ architecture of the proposed ADA-Net is presented. There are five pixel-sampling operations along the channel dimension of selected layers. These samples are stacked together and given as input to five different MLPs as $\Phi = \left\{ \phi_m \right\}^5_{m=1}$ for pixel-wise contrastive learning used in \eqref{eq:cont_loss_pixel}. Frequency domain patches are fed into $\theta$ for patch-wise contrastive learning to computed $\mathcal{L}_\text{freq}$ in \eqref{eq:cont_loss_patch}. The input image is from the source domain USA, while the transformed image appears to have characteristics similar to the Finland domain images.}
    \label{fig:ada_net}
\end{figure*}

Given sets of samples from two different domains: $\left\{\mathbf{x}_i \right\}_{i=1}^{N_x}$ and $\left\{\mathbf{y}_j\right\}_{j=1}^{N_y}$, the generator network learns the transformation $G: X \rightarrow Y$ between the domains $\mathbf{x}_i \in X$, $\mathbf{y}_j \in Y$ where $X, Y \in \mathbb{R}^{P \times N \times C}$. For the sake of simplicity, we will omit $\left\{ i, j \right\}$ indices throughout the remainder of the paper. The architecture of the ADA-Net consisting of multiple sub-networks is presented in Fig \ref{fig:ada_net}. The total loss used for training is expressed as follows,
\begin{equation}
\label{eq:objective}
    \begin{split}
    \mathcal{L}_{\text{T}}\left( G, D, \mathbf{x}, \mathbf{y} \right) & = \mathcal{L}_A \left( G, D, \mathbf{x}, \mathbf{y} \right) \\
    & + \lambda \mathcal{L}_{\text{Spatial}}\left( G, \Phi, \mathbf{x}, \mathbf{y} \right) \\
    & + \beta \mathcal{L}_{\text{IDSpatial}}\left(G, \Phi, \mathbf{y}, \mathbf{x} \right) \\
    & + \gamma \mathcal{L}_{\text{Freq}}\left(G, \theta, \mathbf{x}\right) \\
    & + \vartheta \mathcal{L}_{\text{IDFreq}}\left(G, \theta, \mathbf{y}\right),
    \end{split}
\end{equation}
where $\mathcal{L}_A$ is adversarial, $\mathcal{L}_{\text{Spatial}}$ and $\mathcal{L}_{\text{Freq}}$ are spatial and frequency contrastive learning losses, respectively. Their corresponding identity losses for contrastive learning are $\mathcal{L}_{\text{IDSpatial}}$ and $\mathcal{L}_{\text{IDFreq}}$. While the generator tries to produce synthetic images that are realistic to the target domain, the discriminator learns to distinguish synthetic samples from the real ones in the training set. Given a source domain image sample $\mathbf{x}$, the target domain image is produced by $G\left(\mathbf{x}\right) = \hat{\mathbf{y}}$. Next, the discriminator computes the following 2-D mask $D\left(\hat{\mathbf{y}}\right) = \hat{\mathbf{M}}_y \in \mathbb{R}^{d_p \times d_r}$ representing pixel-wise estimations of real or synthetic decisions. Note that depending on different configurations and downsampling procedures, the discriminator can produce scalar $\hat{m}_y \in \mathbb{R}^1$ decisions as well.

One back-propagation iteration consists of sequential and adversarial training of the generator and discriminators. First, the model weights are updated by $\min_G \mathcal{L}_A\left( G, D, \mathbf{x}, \mathbf{y} \right)$, and then $\max_{D} \mathcal{L}_A\left( G, D, \mathbf{x}, \mathbf{y} \right)$. In this way, the generator network produces images substantially similar to the target domain samples, and the discriminator fails to differentiate real and synthetic images. In the following section, we will detail the training procedure of the ADA-Net, which is divided into two steps.

\begin{figure}[t!]
    \centering
    \includegraphics[width=.97\linewidth]{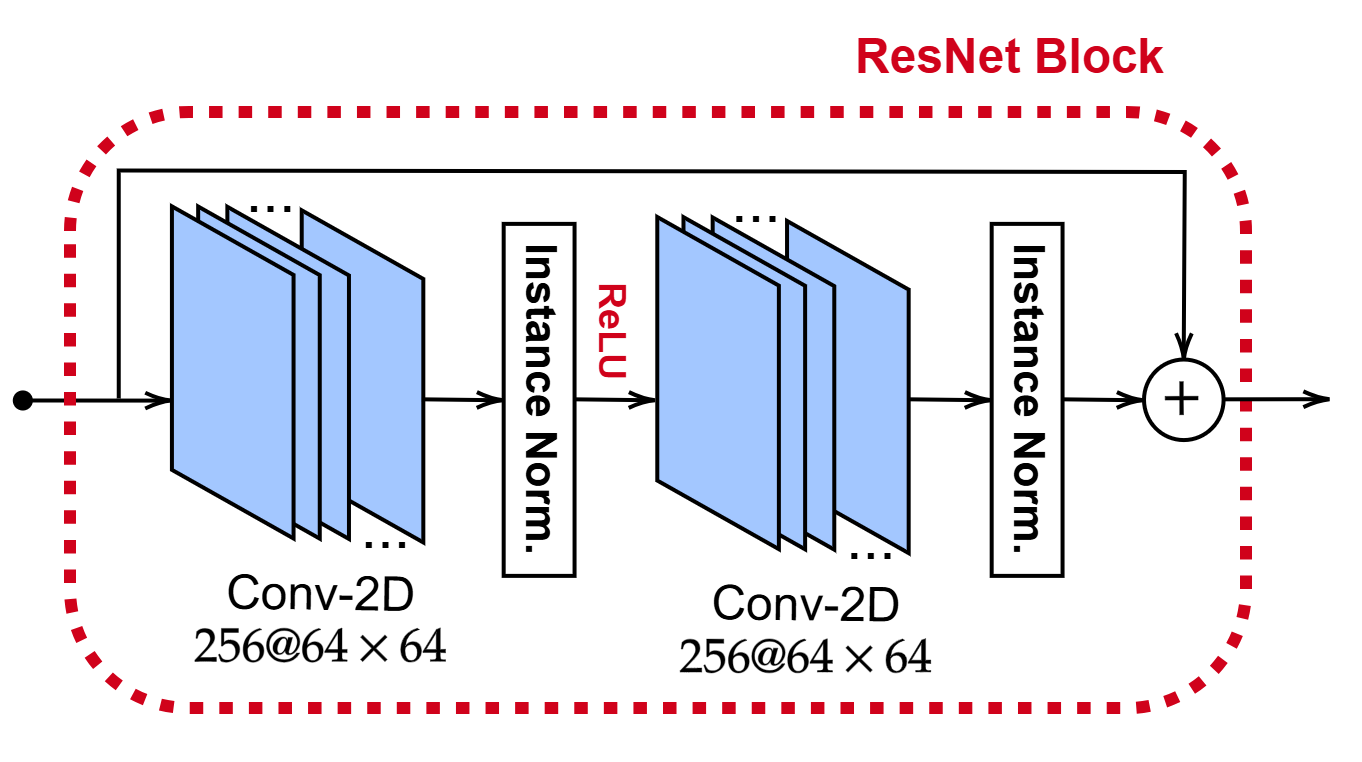}
    \caption{ResNet block is detailed used in the generator network $G$.}
    \label{fig:resnet_block}
\end{figure}

\begin{figure}[t!]
    \centering
    \includegraphics[width=\linewidth]{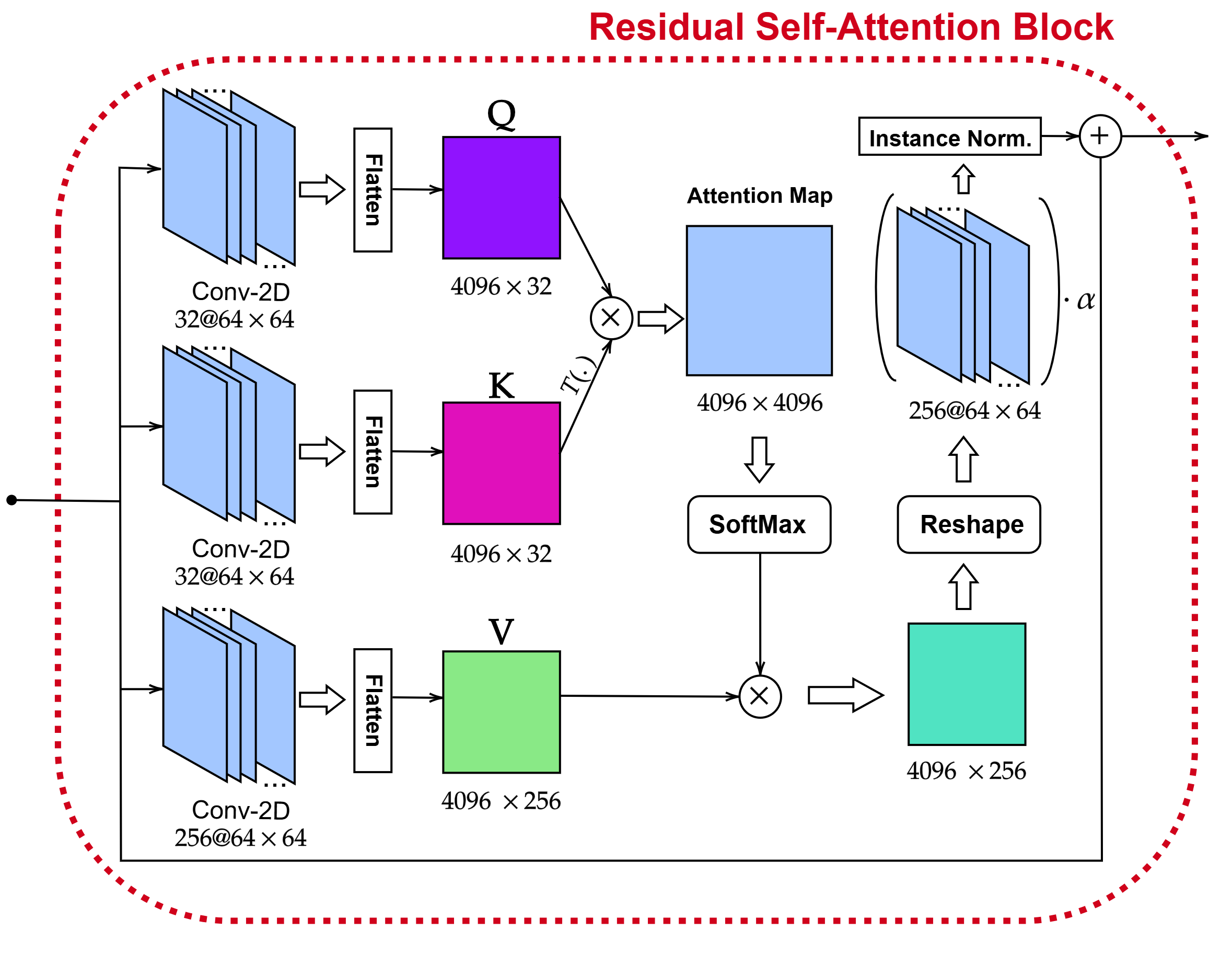}
    \caption{Residual Self-Attention block with convolutional projection operations is used in the generator network $G$, where $\alpha$ is a learnable parameter in \eqref{eq:attention_final}.}
    \label{fig:attention_block}
\end{figure}

\subsubsection{Generator Network Training} The proposed generator network architecture $G$ is illustrated in Fig \ref{fig:ada_net}. All convolutional layers have a kernel size of $5$ except the first and last ones, where the kernel size is set to $7 \times 7$. There are four residual blocks (ResNet \cite{resnet50}) depicted in Fig. \ref{fig:resnet_block}. Furthermore, the generator is equipped with self-attention operations with residual self-attention blocks illustrated in Fig. \ref{fig:attention_block}. A typical multiplicative attention, also called scaled dot product attention \cite{vaswani2017attention} is expressed as,
\begin{equation}
    \label{eq:attention}
    \text{Attention}\left( \mathbf{Q}, \mathbf{K}, \mathbf{V} \right) = \text{SoftMax} \left( \frac{\mathbf{Q} \mathbf{K}^T}{\sqrt{d_q}} \mathbf{V} \right),
\end{equation}
where $\mathbf{Q}, \mathbf{K} \in \mathbb{R}^{d_q}$ and $\mathbf{V} \in \mathbb{R}^{d_v}$. This traditional attention mechanism has certain drawbacks related to gradient vanishing and computational complexity. As highlighted in \cite{vaswani2017attention}, the attention map values may expand substantially due to the multiplication, and they can be located at the edge regions of the SoftMax function where the gradients are considerably smaller. Moreover, $\mathbf{Q}$, $\mathbf{K}$, and $\mathbf{V}$ are generally obtained by fully connected dense layers by linear projection \cite{chen2021crossvit, vaswani2017attention, adalioglu2023saf} causing significant computational complexity and a large number of trainable parameters. While one can use multi-head attention \cite{chen2021crossvit, adalioglu2023saf} to enable parallel computation and enhance computational efficiency, the maps for $\mathbf{Q}$, $\mathbf{K}$, and $\mathbf{V}$ are still computed by linear projection. To address these limitations of traditional attention mechanisms related to gradient vanishing and computational complexity, the linear projection can be replaced by convolutional projection as proposed by \cite{xu2023cur}. Correspondingly, in this study, we propose to use a residual self-attention block with convolutional projection. Let $\mathbf{S} = \left[ \mathbf{S}_1, \mathbf{S}_2, ..., \mathbf{S}_{d_m} \right] \in \mathbb{R}^{d_m \times d_r \times d_c}$ be a given input feature map tensor in an intermediate layer, the convolutional projection operation computing the $j^\text{th}$ feature map component in the next layer is defined as follows,
\begin{equation}
    \label{eq:attention_projection}
        \psi_j\left(\mathbf{S}, \mathbf{W}_j \right) : = \sum_{i=1}^{d_m} \text{conv2D} \left( \mathbf{S}_i, \mathbf{w}_{i, j} \right) + \mathbf{b}_j,
\end{equation}
where $\mathbf{W}_j = [\mathbf{w}_{1, j}, ..., \mathbf{w}_{i, j}, ..., \mathbf{w}_{d_m, j}]$ and $\mathbf{w}_{i, j} \in \mathbb{R}^{f_s \times f_s}$. Overall, the respective representations can be obtained:
\begin{equation}
\label{eq:rep}
    \begin{split}
        \mathbf{Q} = \sum_j \psi_j\left(\mathbf{S}, \mathbf{W}_j^q \right), \\
        \mathbf{K} = \sum_j \psi_j\left(\mathbf{S}, \mathbf{W}_j^k \right), \\
        \mathbf{V} = \sum_j \psi_j\left(\mathbf{S}, \mathbf{W}_j^v \right).
    \end{split}
\end{equation}
Although this projection is more computationally efficient than the traditional attention mechanism, the convolutional operation has greater complexity than, for example, $1 \times 1$ pointwise convolutional projection followed by the study in \cite{zhang2019self}. On the other hand, the proposed approach leverages compact network configurations enabling the usage of 2-D convolutional projection to be feasible and practical within the proposed framework. The residual block then can be written as,
\begin{equation}
\label{eq:attention_final}
        \mathbf{S}_o = \alpha \cdot \text{Attention}\left( \mathbf{Q}, \mathbf{K}, \mathbf{V} \right) + \mathbf{S},
\end{equation}
where the attention operation is provided in \eqref{eq:attention} and $\mathbf{S}$ is the input feature map used to obtain $\mathbf{Q}$, $\mathbf{K}$, $\mathbf{V}$ in \eqref{eq:rep}.

Let $\mathbf{x} \in X$ and $\mathbf{y} \in Y$ be unpaired training samples, when training the generator, adversarial loss in \eqref{eq:objective} is denoted by $\mathcal{L}_{A_G}$ and it can be written as,
\begin{equation}
\label{eq:adv-cost}
    \mathcal{L}_{A_G}\left(G, D, \mathbf{x}\right) = \left \| D\left( G \left( \mathbf{x} \right) \right) - \mathbf{1} \right \|_2^2.
\end{equation}

For contrastive learning, mutual information maximization has been followed \cite{cut, oord2018representation}. Accordingly, contrastive loss is calculated for a given query, its corresponding positive sample, and $K$ number of negative samples in the dataset. In the domain adaptation task, the query is defined as the output $\hat{\mathbf{y}}$, the positive sample is the given input $\mathbf{x}$, and the negative samples are other samples from the source domain $X$. Specifically, the query, positive, and negative samples lie in joint $d-$dimensional joint subspace mapped by $f: \mathbb{R}^{P \times N} \rightarrow \mathbb{R}^d$ where they are denoted as $\mathbf{f}_q, \mathbf{f}_p \in \mathbb{R}^d$, and $\mathbf{F}_n = \left[ \mathbf{f}_{n,1}, ..., \mathbf{f}_{n,K} \right] \in \mathbb{R}^{d \times K}$. Then, the contrastive loss is defined as,
\begin{equation}
\label{eq:cont_loss}
    \mathcal{L}_{C} \left( \mathbf{f}_q, \mathbf{f}_p, \mathbf{F}_n \right) = - \log \left(  \frac{ e^{ \left( \mathbf{f}_q^T \mathbf{f}_p \right) / \tau} }{ e^{ \left( \mathbf{f}_q^T \mathbf{f}_p \right) / \tau} + \sum_{k=1}^K e^{\left( \mathbf{f}_q^T \mathbf{f}_{n, k} \right) / \tau} } \right).
\end{equation}
Note that such a contrastive learning approach has been followed by studies \cite{cut, bachman2019learning, henaff2020data} originating from the idea of Noise Contrastive Estimation (NCE) \cite{gutmann2012noise}, and it has been discussed that patch-wise contrastive learning brings several advantages in preserving structural similarities between transformed images \cite{cut}. This approach tries to cluster nearby patches together in the latent space representation while maximizing the distances between spatially faraway patches. We follow the approach utilized in \cite{cut} by selecting features during the domain transformation of $G$, where these features are then concatenated and fed to multiple Multi-Layer Perceptron (MLP) models. In this case, the mapping of $f: \mathbb{R}^{P \times N} \rightarrow \mathbb{R}^d$ represents the combination of the encoder and several projection networks, where the encoder is trained through the generator.

Accordingly, $M$ layers are selected from different connections of the transformer $G$ network. The query pixel sample features, i.e., $\mathbf{s}_q = \left[ \mathbf{s}_{q, 1}, ..., \mathbf{s}_{q, M} \right] \in \mathbb{R}^{d_m \times M}$ are extracted from these $M$ layers when the transformer network is reintroduced with the produced target domain image as $G\left( \hat{\mathbf{y}} \right)$, where $d_m$ is the feature vector dimension of $m^\text{th}$ layer. On the other hand, positive and negative pixel sample features, denoted as $\mathbf{s}_p = \left[ \mathbf{s}_{p, 1}, ..., \mathbf{s}_{p, M} \right] \in \mathbb{R}^{d_m \times M}$ and $\mathbf{S}_n = \left[ \mathbf{s}_{n, m}^k \right]_{m=1, k=1}^{M, \quad K} \in \mathbb{R}^{d_m \times M \times K}$, respectively, are obtained by feeding the generator with source domain images $G\left( \mathbf{x} \right)$. Overall, $N_s$ is the number of pixel-wise samples extracted along the channel dimension of selected layers. These $M$ sets of $N_s$ feature samples, each denoted as $\mathbf{s}_m \in \mathbb{R}^{d_m}$ having different dimensions, are stacked together and they are given as input to a group of $M$ number of MLPs, i.e. $\Phi = \left\{ \phi_m \right\}^M_{m=1}$. Each extracted sample is normalized to have unit $\ell_2-$norm. The learned subspace representation of the query, positive, and negative samples are then obtained as $\mathbf{f} = \left\{ \phi_1\left( \mathbf{s}_1 \right), \phi_2\left( \mathbf{s}_2 \right), ..., \phi_M\left( \mathbf{s}_M \right) \right\}$. Correspondingly, the pixel-wise spatial contrastive loss is obtained by averaging as follows,
\begin{equation}
\label{eq:cont_loss_pixel}
    \mathcal{L}_{\text{C}}^* = \frac{1}{M}\sum_{m=1}^M \mathcal{L}_{C} \left( \phi_m\left( \mathbf{s}_{q, m} \right), \phi_m\left( \mathbf{s}_{p, m} \right), \left[ \phi_m\left( \mathbf{s}_{n, m}^k \right) \right]_{k=1}^K \right),
\end{equation}
where $\mathbf{s}_{q, m}, \mathbf{s}_{p, m} \in \mathbb{R}^{d_m}$, and $\mathbf{S}_{n, m} = \left[ \mathbf{s}_{n, m}^1, ..., \mathbf{s}_{n, m}^K \right] \in \mathbb{R}^{d_m \times K}$. Then, this is further accumulated by going through the selected $N_s$ number of pixel samples by alternating the following triplet \cite{cut}: $\left\{ \mathbf{s}_q, \mathbf{s}_p, \mathbf{S}_n \right\}$; assuming $i \in \left\{1, 2, ..., N_s \right\}$ is the index operator for selected sample set, and total spatial contrastive loss is computed by 
\begin{equation}
\label{eq:cont_loss_total}
    \mathcal{L}_{\text{Spatial}} = \frac{1}{N_s} \sum_{i=1}^{N_s} \mathcal{L}_{\text{C}}^* \left( \Phi, \mathbf{s}_{q}^i, \mathbf{s}_{p}^i, \mathbf{S}_{n}^{N_s \setminus i} \right),
\end{equation}
which follows that $K = N_s - 1$.

In the framework, we use three-layer MLPs as depicted in Fig \ref{fig:ada_net}. As followed by the study \cite{cut}, there is $256$ number of neurons in the hidden and output layers corresponding to have $d_m = 256$, i.e., the subspace representation dimensionality of triplets. The number of patches is $256$, i.e., $K=N_s - 1 = 255$, and the number of selected layers in the generator where the pixel-wise samples are taken is set to $M = 5$.

Identity spatial contrastive loss $\mathcal{L}_{\text{IDSpatial}}$ is computed using $\left\{ \Tilde{\mathbf{s}}_q, \Tilde{\mathbf{s}}_p, \Tilde{\mathbf{S}}_n \right\}$ triplets that are obtained when target domain images $\mathbf{y} \in Y$ are introduced to the generator network. Accordingly, let $G\left( \mathbf{y} \right) = \Tilde{\mathbf{y}}$ is obtained, the aim is to preserve the input image with minimal changes after the transformation since it already belongs to the target domain. Therefore, $\Tilde{\mathbf{s}}_q$ is extracted from $G\left( \Tilde{\mathbf{y}} \right)$ while $\Tilde{\mathbf{s}}_p$ and $\Tilde{\mathbf{S}}_n$ are obtained by $G\left( \mathbf{y} \right)$. Then, identity spatial loss is defined as $\mathcal{L}_{\text{IDSpatial}} = \frac{1}{N_s} \sum_{i=1}^{N_s} \mathcal{L}_{\text{C}}^* \left( \Phi, \Tilde{\mathbf{s}}_{q}^i, \Tilde{\mathbf{s}}_{p}^i, \Tilde{\mathbf{S}}_{n}^{N_s \setminus i} \right)$.

Certain patterns of different images can be revealed in the frequency domain, which is often overlooked in the spatial domain. Thus, leveraging the frequency domain representation of the images during the training of GAN models can be preferable, as discussed in \cite{jiang2021focal}. To this end, we propose to use representation coefficients of the images in the frequency domain for patch-wise contrastive loss computation. Accordingly, $\mathcal{L}_{\text{freq}}$ in \eqref{eq:objective} is defined as,
\begin{equation}
\label{eq:cont_loss_patch}
    \mathcal{L}_{\text{freq}} = \frac{1}{N_f} \sum_{i=1}^{N_f} \mathcal{L}_{C} \left( \theta\left( \mathbf{z}_q^i \right), \theta\left( \mathbf{z}_p^i \right), \left[ \theta\left( \mathbf{z}_n^{k, N_f \setminus i} \right) \right]_{k=1}^{K_f} \right),
\end{equation}
where $\theta$ is an MLP model providing the mapping of $\theta: \mathbb{C}^{d_f \times d_f} \rightarrow \mathbb{R}^d$. The input layer of $\theta$ concatenates real and imaginary parts of a given complex domain sample; then, components are jointly processed through the MLP network. This procedure is illustrated in Fig \ref{fig:ada_net}. The MLP network $\theta$ has a compact three-layer structure with $1024$ and $256$ number of neurons in the hidden and output layers, respectively. The number of extracted patches is $N_f = 64$, and the patch size is $32 \times 32$.

The frequency domain flattened patches are $\mathbf{z}_q^i, \mathbf{z}_p^i \in \mathbb{C}^{d_f}$, and $\mathbf{Z}_n^{N_f \setminus i} = \left[ \mathbf{z}_n^{1, N_f \setminus i}, \mathbf{z}_n^{2, N_f \setminus i}, ..., \mathbf{z}_n^{K_f, N_f \setminus i} \right]\in \mathbb{C}^{d_f \times K_f}$ with $d_f = 256$ and $K_f = N_f - 1 = 63$. These frequency patch samples are obtained through 2D-DFT operation:
\begin{equation}
    \begin{split}
    \mathbf{z}_q^i & = \mathcal{F}\mathcal{P}\left( \mathbf{x}, i \right) \mathcal{F}^T, \\
    \mathbf{z}_p^i & =  \mathcal{F}\mathcal{P}\left( \hat{\mathbf{y}}, i \right) \mathcal{F}^T, \\
    \mathbf{Z}_n^{N_f \setminus i} & = \mathcal{F}\mathcal{P}\left( \mathbf{x}, N_f \setminus i \right) \mathcal{F}^T,
    \end{split}
\end{equation}
where $\mathcal{P}$ is the patch sampler operator for a given index $i \in \left\{1, 2, ..., N_f \right\}$ and $\mathcal{F}$ is Fourier matrix. Similarly, identity frequency loss is written as $\mathcal{L}_{\text{IDFreq}} = \frac{1}{N_s} \sum_{i=1}^{N_s} \mathcal{L}_{C} \left( \theta\left( \Tilde{\mathbf{z}}_q^i \right), \theta\left( \Tilde{\mathbf{z}}_p^i \right), \left[ \theta\left( \Tilde{\mathbf{z}}_n^{k, N_f \setminus i} \right) \right]_{k=1}^{K_f} \right)$, where $\Tilde{\mathbf{z}}_q$ is sampled by $G\left( \Tilde{\mathbf{y}} \right)$, and $\Tilde{\mathbf{z}}_p$ and $\Tilde{\mathbf{Z}}_n$ are by $G\left( \mathbf{y} \right)$. The ADA-Net approach combines spatial and frequency identity losses acting as a regularization mechanism to prevent the network from overreacting to slight changes in both domains.

In particular, we improve two important aspects of the contrastive loss computation. Although the contrastive loss proposed by \cite{cut} is introduced as patch-wise loss, the given input $\mathbf{s}_m \in \mathbb{R}^{d_m}$ to set of MLP networks are pixel-wise collected features from different layers of the generate $G$. Possibly, when the features propagate through the network, the correlation between neighboring features has been lost. Therefore, in this study, the ADA-Net utilizes the patches extracted directly from the images in the frequency domain, accomplishing the actual patch-wise contrastive learning. Secondly, the additional frequency domain information enhances overall structural preservation \cite{jiang2021focal} since certain patterns are more predominantly represented.

\subsection{Discriminator Training}

Discriminator training of the proposed ADA-Net approach is similar to traditional GANs using the following loss:
\begin{equation}
\label{eq:dis_loss}
    \mathcal{L}_{A_D}\left( G, D, \mathbf{x}, \mathbf{y} \right) = \left\| D\left( \mathbf{y} \right) - \mathbf{1} \right\|_2^2 + \left\| D\left( G\left( \mathbf{x} \right) \right)\right\|_2^2
\end{equation}
Minimization of \eqref{eq:dis_loss} is equivalent to $\max_{D} \mathcal{L}_A\left( G, D, \mathbf{x}, \mathbf{y} \right)$. In particular, during a single training iteration, the generator network weights are first updated minimizing \eqref{eq:adv-cost}. This update tricks the discriminator into outputting the mask of ones when fake images are provided. Next, the discriminator network weights are updated minimizing \eqref{eq:dis_loss} for a given set of synthetic and real target domain images $\left\{\hat{\mathbf{y}}, \mathbf{y} \right\}$ where $\hat{\mathbf{y}} = G\left( \mathbf{x} \right)$. Four different network configurations are used as the discriminator. PatchGANs \cite{isola2017image} have $6$ convolutional layers with $4 \times 4$ filter sizes. PixelGANs can be considered as a variation of PatchGANs with a size of $1\times1$ filter sizes. This is a fairly compact discriminator with only $3$ convolutional layers. Next, StyleGAN2 \cite{stylegan} networks are used, including $20$ convolutional layers, $6$ number of ResNet blocks, and $2$ linear layer at the output. Finally, Tile-StyleGAN2 is deployed, which has a similar architecture to StyleGAN2, but the input image of the discriminator has been divided into smaller overlapped image tiles, and each tile has been processed by the network. For a more detailed discussion about the StyleGAN2 discriminators, readers are referred to \cite{stylegan}. Overall, these discriminators are fairly deep networks with advanced normalization techniques consisting of kernel weight modulation (normalization of the weights) and demodulation (ensuring stable standard deviation after the activations). In all discriminator configurations, activation functions have been set to leaky ReLU except the output layer.

\subsection{Flair U-Net Dead Tree Segmentation Approach}

After applying domain adaptation to transform source domain $X$ images to the target domain $Y$ images, we have used the Flair U-Net segmentation model \cite{flair_unet}. This model is a quite deep network in an auto-encoder structure consisting of encoder and decoder parts. It has several skip connections in a U-Net like architecture connecting encoder layers to the decoder. The encoder part has a total number of $16$ ResNet blocks involving several downsampling operations. The decoder is fairly more compact with three gated attention layers and one convolutional layer. In total, the network has $35$ convolutional layers. Authors are referred to \cite{flair_unet} for a more detailed discussion regarding the Flair U-Net model. In practice, training such a deep network requires a tremendous amount of annotated data, whereas the annotated data in the source domain is limited. Therefore, the Flair U-Net model is trained with target domain $Y$ images and their labels, then, it is used to evaluate domain-adapted $X2Y$ images. Hence, the proposed segmentation approach in this study can be considered as a \textit{zero-shot} approach since the source domain images and labels are not seen by the segmentation Flair U-Net model during training.

\subsection{Competing Methods}

The following sections will briefly present several selected Image-to-Image translation approaches. These approaches are utilized in the proposed zero-shot standing dead tree segmentation framework based on performing blind domain adaptation.

\subsubsection{Contrastive Unpaired Translation (CUT)}

The main motivation of the CUT approach \cite{cut} is that there is an existing correlation between patches from the output of the network and input images. Therefore, maximizing mutual information for these correlated locations should increase the Image-to-Image translation performance. This has been accomplished by reducing the distances of patch representations taken from the same locations of input and output while increasing the distance for the different location patches during training. The generator training loss of the CUT model is given as,
\begin{equation}
\label{eq:objective_cut}
    \begin{split}
    \mathcal{L}_{\text{CUT}}\left( G, D, \mathbf{x}, \mathbf{y} \right) & = \mathcal{L}_A\left( G, D, \mathbf{x}, \mathbf{y} \right) + \lambda \mathcal{L}_{\text{Spatial}}\left( G, \Phi, \mathbf{x}, \mathbf{y} \right) \\
    & + \beta \mathcal{L}_{\text{IDSpatial}}\left( G, \Phi, \mathbf{y}, \mathbf{x} \right).
    \end{split}
\end{equation}
Note that this loss function corresponds to the first part of \eqref{eq:objective} excluding patch-wise frequency contrastive learning components. However, as previously mentioned, we argue that the minimized loss function in \eqref{eq:objective_cut} does not fully achieve patch-wise contrastive learning objective since there is no guaranteed correlation between taken pixel samples from the encoder's feature maps.

The generator network of the CUT model consists of a substantial number of $9$ ResNet blocks, which are composed of two convolutional layers each, as shown in Fig. \ref{fig:resnet_block}. Including additional layers, there are a total number of $24$ convolutional layers. In contrast, ADA-Net benefits from the introduced self-attention layers, and even then, it is significantly computationally efficient with few layers and smaller number of trainable parameters compared to the CUT model. Both the CUT and ADA-Net approach use $5$ different MLP models for further projection network in contrastive learning as detailed in Section \ref{sec:ada-net}. Locations of these MLP connections are evenly distributed throughout the generator layers ensuring learning and utilization of different feature-level representations across the generator. Discriminator training of the CUT model is traditional and pretty much follows similar steps in conventional vanilla GANs.

\subsubsection{Fast Contrastive Unpaired Translation (FastCUT)}

FastCUT method \cite{cut} is based on a less computationally demanding version of the CUT method without the identity contrastive learning. While they both share the same network configuration, the loss function used for the training of FastCUT is written as,
\begin{equation}
\label{eq:objective_fastcut}
    \mathcal{L}_{\text{FastCUT}}\left( G, D, \mathbf{x}, \mathbf{y} \right) = \mathcal{L}_A\left( G, D, \mathbf{x}, \mathbf{y} \right) + \lambda \mathcal{L}_{\text{Spatial}}\left( G, \Phi, \mathbf{x}, \mathbf{y} \right).
\end{equation}
In this way, the need for identity mapping is removed when target domain image is given as input to the network together with their feature representation mappings for contrastive learning. It is discussed in \cite{cut} that FastCUT model can achieve comparable performance levels with significantly improved efficiency, e.g., compared to traditional approaches such as Cycle-GANs.

\subsubsection{Cycle-Consistent Generative Adversarial Networks (Cycle-GANs)}

Cycle-GANs are composed of four different networks: two generators providing forward and inverse transformations between the source and target domains: $G: X \rightarrow Y$ and $F: Y \rightarrow X$; and their corresponding two discriminators learning to distinguish the generated images from the real ones producing the following masks: $D_Y\left( \hat{\mathbf{y}} \right) = \hat{\mathbf{M}}_y \in \mathbb{R}^{d_p \times d_r}$ and $D_X\left( \hat{\mathbf{x}} \right) = \hat{\mathbf{M}}_x \in \mathbb{R}^{d_p \times d_r}$. The networks are trained using below overall loss:
\begin{equation}
\label{eq:objective_cyclegan}
    \begin{split}
    & \mathcal{L}_{\text{CycleGAN}}\left( G, F, D_X, D_Y, \mathbf{x}, \mathbf{y} \right) = \mathcal{L}_A\left( G, D_Y, \mathbf{x}, \mathbf{y} \right) \\
    & + \mathcal{L}_A\left( F, D_X, \mathbf{y}, \mathbf{x} \right) + \alpha \mathcal{L}_{\text{cyc}}\left( G, F, \mathbf{x}, \mathbf{y} \right) + \delta \mathcal{L}_{\text{ID}}\left( G, F, \mathbf{x}, \mathbf{y} \right),
    \end{split}
\end{equation}
where $\mathcal{L}_A$, $\mathcal{L}_\text{cyc}$, and $\mathcal{L}_\text{ID}$ are adversarial, cycle-consistency, and identity losses, respectively. The training procedure of the discriminators is pretty much similar to conventional GANs with the addition of a second iteration in back-propagation for the training of two discriminators. Hence, only training procedure of the generators will be briefly discussed in the following. For a given un-paired sample $\left\{ \mathbf{x}, \mathbf{y} \right\}$, minimized adversarial loss for two generators is written as,
\begin{equation}
    \mathcal{L}_{A_{G, F}} = \left \| D_Y\left( G \left( \mathbf{x} \right) \right) - \mathbf{1} \right \|_2^2 + \left \| D_X\left( F \left( \mathbf{y} \right) \right) - \mathbf{1} \right \|_2^2.
\end{equation}
In back-propagation iteration, two cycles are computed $F\left(G\left( \mathbf{x} \right) \right): \mathbf{x} \rightarrow \hat{\mathbf{y}} \rightarrow \tilde{\mathbf{x}}$ and $G\left(F\left( \mathbf{y} \right) \right): \mathbf{y} \rightarrow \hat{\mathbf{x}} \rightarrow \tilde{\mathbf{y}}$ in order to compute cycle-consistency loss:
\begin{equation}
\label{eq:cyc-cost}
    \begin{split}
    \mathcal{L}_{\text{cyc}}\left( G, F, \mathbf{x}, \mathbf{y} \right) = \left \| F\left(G\left( \mathbf{x} \right) \right) - \mathbf{x} \right \|_1 + \left \| F\left(G\left( \mathbf{y} \right) \right) - \mathbf{y} \right \|_1.
    \end{split}
\end{equation}
Lastly, the identity loss $\mathcal{L}_\text{Id}$ penalizes the changes at the output for the generator networks when their corresponding domain samples are introduced to the forward and inverse mappings:
\begin{equation}
    \mathcal{L}_{\text{ID}}\left( G, F, \mathbf{x}, \mathbf{y} \right) = \left \| G\left( \mathbf{y} \right) - \mathbf{y} \right \|_1 + \left \| F\left( \mathbf{x} \right) - \mathbf{x} \right \|_1.
\end{equation}
For both forward and inverse generators $G$ and $F$, the same network configuration with the CUT network including $9$ ResNet blocks \ref{fig:resnet_block} are used consisting of a total number of $24$ convolutional layers.

In an Image-to-Image translation task, the aim is to preserve mutual information. Cycle-GANs promote this by generating back input images from produced images through complete cycles. Thus, it is ensured that certain image characteristics are maintained during the transformations with both generators. While training such a framework consisting of multiple iterations and sub-networks is computationally expensive, Cycle-GANs offer an alternative approach to the CUT model which utilizes mutual information maximization via contrastive learning.

\section{Experimental Evaluation}
\label{sec:experimental_results}

In this work, one of the objectives was to explore the feasibility of using Image-to-Image translation methods for blind domain adaption, particularly for standing dead tree segmentation from aerial images. An extensive set of experiments has been conducted in order to evaluate the proposed ADA-Net approach against the existing methods and to investigate the first objective.

\subsection{Experimental Setup}

\begin{figure}[h]
    \centering
    \includegraphics[width=1\linewidth]{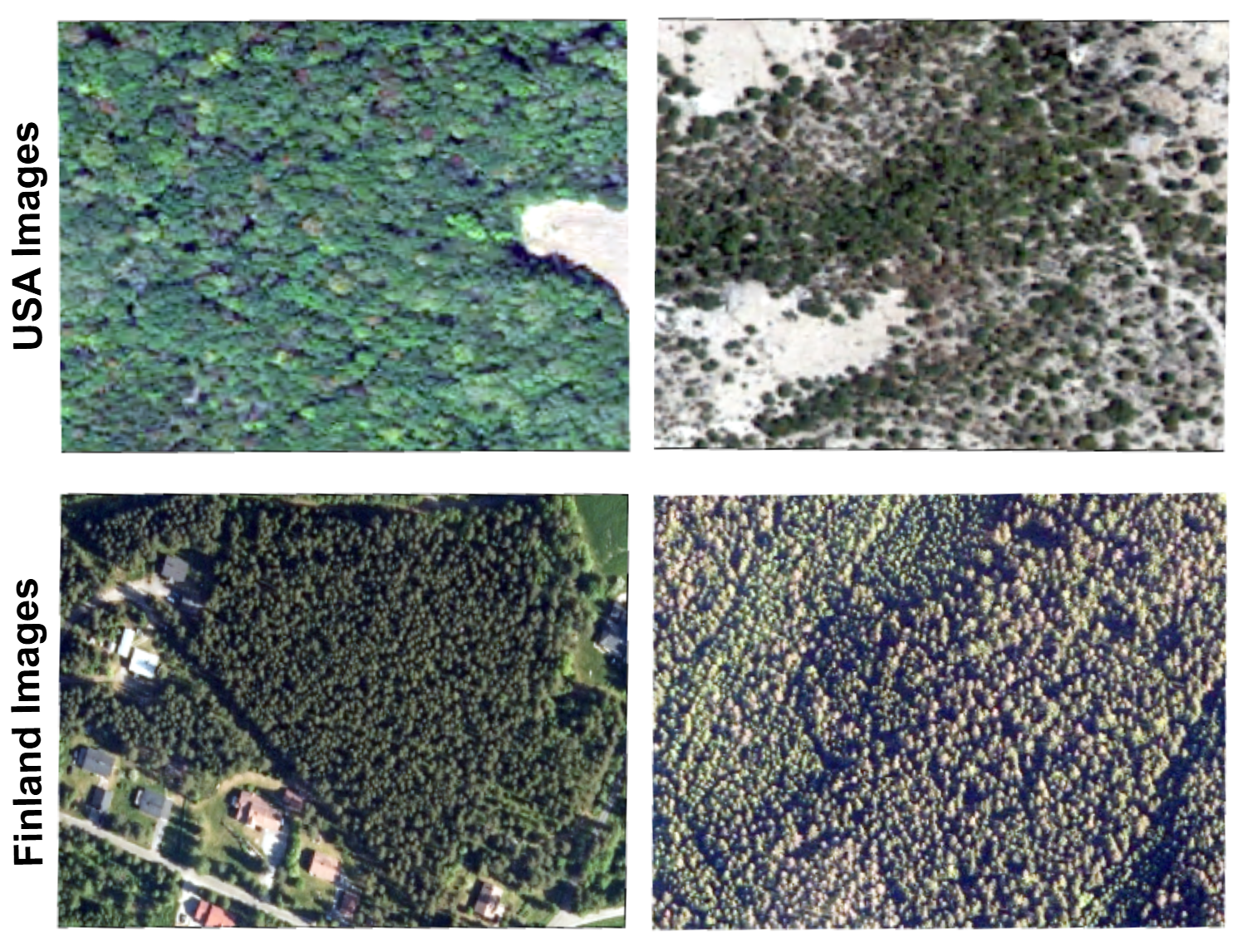}
    \caption{Sample images are provided from the USA and Finland datasets in RGB illustration.}
\label{fig:dataset}
\end{figure}

Target and source domain datasets consist of multispectral aerial images from Finland and the USA, respectively. The dataset from Finland consists of openly available aerial orthoimagery from the years 2011-2023 provided by the National Land Survey of Finland \cite{finland}. The images have a ground sampling distance of $0.5$ meters, but have been resampled to $0.25$ m resolution. The dataset from the USA consists of openly available aerial orthophotos from the years 2018-2023 collected for the National Agriculture Imagery Program \cite{naip}. The imagery ranges multiple states and has a ground sampling distance and resolution of $0.6$ m. Sample images from two sites are illustrated in Fig. \ref{fig:dataset}. It appears that the domain images are significantly different from each other, i.e., the USA images tend to be brighter, while the Finland images have more occluded regions and shadows. Furthermore, tree species from two distinct biomes are also substantially dissimilar. Accordingly, the data have four channels, including R, G, B, and near-infrared (NIR). Utilizing such a multi-view approach with the NIR channel has several advantages, e.g., increased contrast between natural and non-vegetation iteration and existing spectral trace of dead vegetation at the NIR band. Therefore, we aim to increase the dead tree segmentation performance by leveraging four-channel data.

From the Finland site, there are $124$ scenes available with image dimensions averaging greater than 5,000 × 5,000 pixels. We extracted overlapped image tiles with a size of $256 \times 256$ using a stride size of $128$ from these scenes. After preprocessing, there are a total number of $87,539$ image tiles available. It is desired to have equally represented dead trees in the image scenes alongside those without any dead trees. Hence, in the segmentation dataset, $26768$ training, $8104$ validation, and $4810$ testing image tiles are randomly selected to construct a class-balanced dataset from the total number of tiles. Note the fact that such a procedure is followed only for the Finland data since the USA data is already balanced in nature. In our objective of experimental evaluations, we focus on utilizing the target domain $Y$, Finland images in order to train the segmentation Flair U-Net model. Then, we transform USA images into the Finland domain using the proposed domain adaptation approach.

In the USA data, there are $444$ annotated scenes available with relatively smaller dimensions around $300 \times 300$ pixels. The data are manually annotated by our collaborator group of forest health experts. The annotation procedure is summarized as follows: (i) a random $3 \times 3$ km region is selected, (ii) an initial inspection is carried out over the selected region, (iii) a rectangular sub-region possible to contain dead trees is identified for further evaluation, and (iv) the sub-region is manually labelled, cropped and saved as the annotated scene. These scenes are further split into $70\%$ (312 scenes), $10\%$ (44 scenes), and $20\%$ (88 scenes), for train, validation, and test, respectively. Then, we extract image tiles from the train split with a size of $256 \times 256$ using a stride size of $64$ for the USA data. The resulting number of training image tiles is $2346$. We randomly select another $2346$ image tiles from Finland training split to train the domain adaptation networks. Overall, the networks trained to generate \textit{USA2Finland} transformation have used a total number of $2346 + 2346 = 4728$ images with $256 \times 256$ image dimensions. It is worth mentioning that the validation and test splits of the USA data have been inferred without extracting overlapping patches for the fair segmentation evaluation.

Experiments have been performed using PyTorch \cite{paszke2019pytorch} library in Python. Reported computational complexity evaluations are carried out on a computing node having AMD 7H12 CPU and NVIDIA® A100 GPU with 128 GB of system memory. We use the ADAM optimizer \cite{kingma2014adam} with the momentum update decay rates set to $\beta_1 = 0.5$ and $\beta_2 = 0.999$. The proposed approach and competing method networks have all been trained for $60$ epochs using a batch size of $8$, and the best weights are selected according to the epoch number giving the minimum validation loss. The hyperparameters of the loss functions are chosen empirically and they are provided in Table \ref{tab:hyperparameters}. The learning rate is selected as $2\times10^{-6}$ for the proposed ADA-Net approach, $2\times10^{-5}$ for CUT and Cycle-GAN methods, and as $2\times10^{-4}$ for FastCUT method.

\begin{table}[]
\centering
\caption{Emprically selected hyperparameter values \\ for the methods used in this study.}
\label{tab:hyperparameters}
\resizebox{0.9\linewidth}{!}{
 \begin{tabular}{@{}ccc@{}}
\toprule
\textbf{Hyperparameter}          & \textbf{Value} & \textbf{Method} \\ \midrule
$\lambda$ in Eqs. \eqref{eq:objective}, \eqref{eq:objective_cut} & $0.5$ & ADA-Net, CUT \\
$\lambda$ in Eq. \eqref{eq:objective_fastcut} & $10$ & FastCUT \\
$\beta$ in Eqs. \eqref{eq:objective}, \eqref{eq:objective_cut} & $0.5$ & ADA-Net, CUT \\
$\gamma$, $\vartheta$ in Eq. \eqref{eq:objective} & $0.5$ & ADA-Net \\
$\tau$ in Eq. \eqref{eq:cont_loss} & $0.07$ & ADA-Net, CUT, FastCUT \\
$\alpha$ in Eq. \eqref{eq:objective_cyclegan} & $10$ & Cycle-GAN \\
$\delta$ in Eq. \eqref{eq:objective_cyclegan} & $5$ & Cycle-GAN \\ \bottomrule
\end{tabular}
}
\end{table}

\subsection{Results}

\begin{figure*}[!h]
    \centering
    \includegraphics[width=1\linewidth]{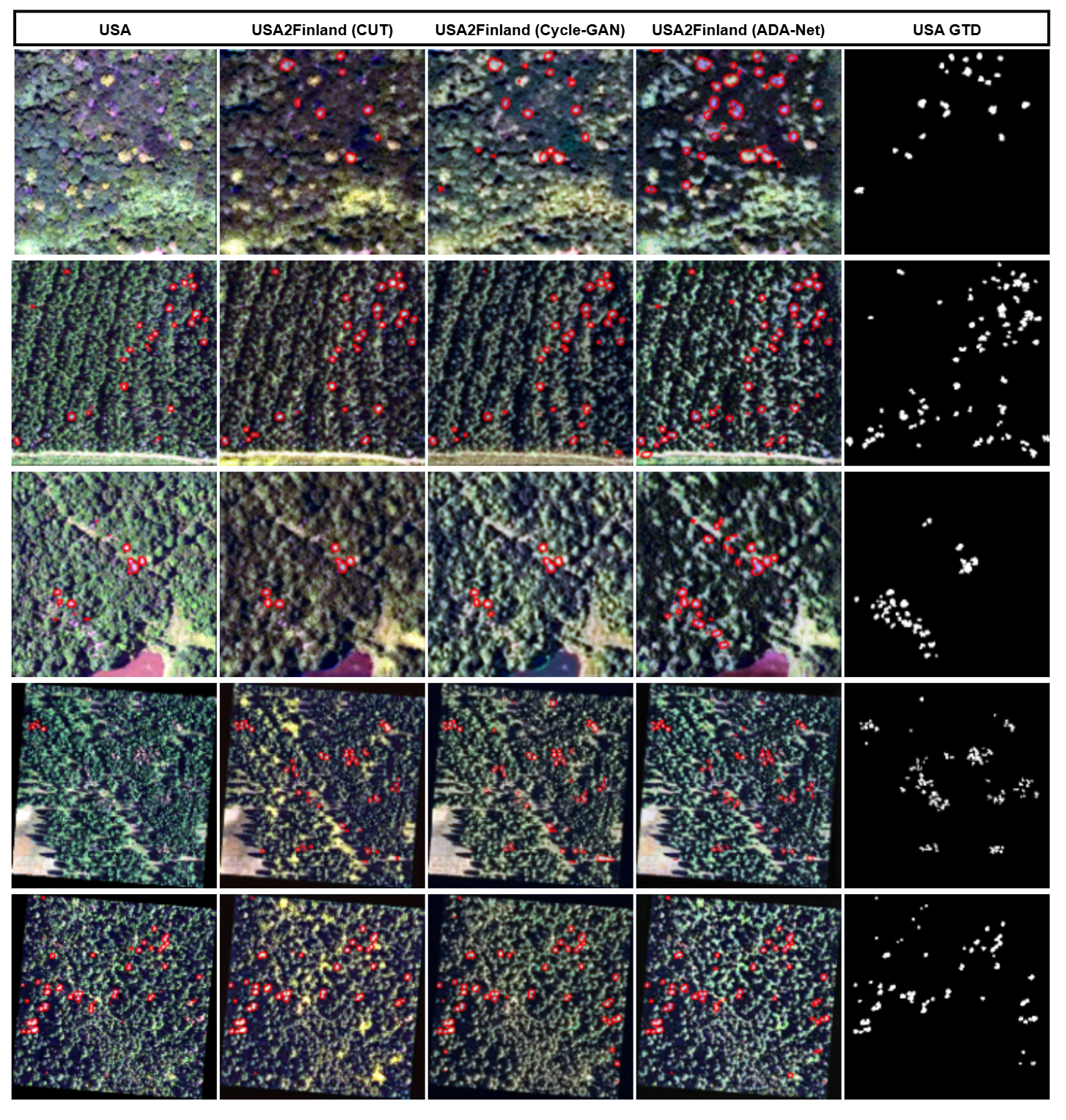}
    \caption{Predicted standing dead trees by Flair U-Net model over USA2Finland images after applying domain adaptation with the proposed ADA-Net approach and two best competitors including CUT and Cycle-GAN approaches. The first column results are obtained without domain adaptation, and the last column indicates the ground truth data (GTD) used for the performance evaluations.}
\label{fig:segmentation_results}
\end{figure*}

In quantitative evaluation, we consider the standing dead tree segmentation task using the transformed images. Accordingly, given the number of true positive (TP), true negative (TN), false positive (FP), and false negative (FN) pixels computed between the true and predicted dead tree segmentation masks, the following metrics are used in the segmentation performance evaluation after the domain adaptation is performed:
\begin{equation}
    \text{Dice Score} = \frac{2 \times \text{TP}}{(2 \times \text{TP} + \text{FP} + \text{FN})},
\end{equation}
\begin{equation}
    \text{Accuracy} = \frac{\text{TP} + \text{TN}}{\text{TP} + \text{TN} + \text{FP} + \text{FN}},
\end{equation}
\begin{equation}
    \text{Specificity} = \text{TN} / (\text{TN} + \text{FP}),
\end{equation}
and we use $F_2-\text{Score}$ since the dice score metric already shows $F_1-$Score characteristics and $F_2-$Score has more tendency to prioritize higher sensitivity levels. Accordingly, it is computed with $a = 2$,
\begin{equation}
    F_a = (1 + a ^ 2) \frac{\text{Precision} \times \text{Sensitivity}}{a ^ 2 \times \text{Precision} + \text{Sensitivity}},
\end{equation}
where precision and sensitivity are obtained as,
\begin{equation}
    \text{Precision} = \text{TP} / (\text{TP} + \text{FP}),
\end{equation}
\begin{equation}
    \text{Sensitivity} = \text{TP} / (\text{TP} + \text{FN}).
\end{equation}
Finally, Intersection over Union (IoU) is computed as,
\begin{equation}
    \text{IoU} = \text{TP} / (\text{TP} + \text{FP} + \text{FN}).
\end{equation}

The dataset of annotated aerial optical images from the Finland site contains a significantly large number of images, and therefore, they are suitable for training a deep network for standing dead tree segmentation. On the other hand, only $2346$ extracted image tiles are available for the USA site. The Flair U-Net model is trained using annotations with Finland data, and the trained network is then transferred to segment dead trees in the USA without fine-tuning. The segmentation results are given in Table \ref{tab:results} when the USA images are adapted to the Finland domain. We name these newly generated images as \textit{USA2Finland} images. Note that since no segmentation annotations are used during the training of domain adaptation networks from the USA images, and the Flair-UNet segmentation model is transferred without fine-tuning, one can say that the proposed procedure in this study to segment USA region standing dead trees is a zero-shot segmentation.

\begin{table*}[ht!]
\caption{Obtained standing dead tree segmentation performance levels are provided over USA2Finland, where the USA images are transferred/adapted into Finland domain images using the proposed and compared approaches with different discriminator configurations. Accordingly, the Flair U-Net segmentation model is applied to the transformed images. The best and second best results are highlighted in red and blue.}

\centering
\label{tab:results}
\resizebox{\linewidth}{!}{
\begin{tabular}{@{}ccccccccc@{}}
\toprule
\textbf{Method} & \textbf{Discriminator} & \textbf{Dice Score} & \boldmath$F_2-$\textbf{Score} & \textbf{IoU} & \textbf{Accuracy} & \textbf{Precision} & \textbf{Specificity} & \textbf{Sensitivity} \\ 
\midrule
\multicolumn{1}{c}{\multirow{6}{*}{\rotatebox[origin=c]{0}{  CUT \cite{cut}}}} \vspace{0.15cm} & PatchGAN \cite{isola2017image} & $0.3610$ & $0.3046$ & $0.2202$ & $0.9836$ & $0.5217$ & $0.9957$ & $0.2759$ \\

\vspace{0.15cm} & PixelGAN & $0.3772$ & $0.3081$ &	$0.2324$ & \textcolor{blue}{\boldmath$0.9848$} & $0.6026$ & $0.9969$ & $0.2745$ \\

\vspace{0.15cm} & StyleGAN2 \cite{stylegan} & $0.3460$ & $0.2975$ & $0.2092$ &	$0.9827$ & $0.4749$ & $0.9949$ & $0.2721$ \\

\vspace{0.1cm} & Tile-StyleGAN2 \cite{stylegan} & $0.3927$ & $0.3288$ & $0.2443$ & $0.9846$ & $0.5805$ & $0.9963$ & $0.2967$ \\ \midrule

\multicolumn{1}{c}{\multirow{6}{*}{\rotatebox[origin=c]{0}{ FastCUT \cite{cut}}}}  \vspace{0.15cm} & PatchGAN \cite{isola2017image} & $0.2870$ & $0.2158$ & $0.1676$ & $0.9845$ & $0.6375$ & $0.9982$ & $0.1852$ \\

\vspace{0.15cm} & PixelGAN & $0.3062$ & $0.2345$ & $0.1808$ & $0.9845$ & $0.6244$ & $0.9979$ & $0.2029$ \\

\vspace{0.15cm} & StyleGAN2 \cite{stylegan} & $0.3614$ & $0.3109$ & $0.2206$ & $0.9831$ & $0.4955$ & $0.9950$ & $0.2844$ \\

\vspace{0.1cm} & Tile-StyleGAN2 \cite{stylegan} & $0.2759$ & $0.2017$ & $0.1600$ & \textcolor{red}{\boldmath$0.9849$} & \textcolor{red}{\boldmath$0.7135$} & \textcolor{red}{\boldmath$0.9988$} & $0.1710$ \\ \midrule

\multicolumn{1}{c}{\multirow{6}{*}{\rotatebox[origin=c]{0}{ Cycle-GAN \cite{cycle_gan}}}}  \vspace{0.15cm} & PatchGAN \cite{isola2017image} & $0.3759$ & $0.3322$ & $0.2314$ & $0.9828$ & $0.4815$ & $0.9943$ & $0.3083$ \\

\vspace{0.15cm} & PixelGAN & $0.3736$ & $0.3110$ & $0.2297$ & $0.9842$ & $0.5625$ & $0.9963$ & $0.2797$ \\

\vspace{0.15cm} & StyleGAN2 \cite{stylegan} & $0.3571$ & $0.3358$ & $0.2174$ & $0.9805$ & $0.3995$ & $0.9917$ & $0.3229$ \\

\vspace{0.1cm} & Tile-StyleGAN2 \cite{stylegan} & $0.3754$ & $0.3735$ & $0.2311$ & $0.9792$ & $0.3787$ & $0.9896$ & \textcolor{blue}{\boldmath$0.3722$} \\ \midrule

\multicolumn{1}{c}{\multirow{6}{*}{\rotatebox[origin=c]{0}{\makecell{\textbf{ADA-Net} \\ \textbf{(Proposed)}}}}} \vspace{0.15cm} & PatchGAN \cite{isola2017image} & \textcolor{blue}{\boldmath$0.4232$} & $0.3710$ & \textcolor{blue}{\boldmath$0.2684$} & $0.9843$ &	$0.5531$ & $0.9953$ & $0.3428$ \\

\vspace{0.15cm} & Pixel & $0.3057$ & $0.2315$ & $0.1804$ & \textcolor{blue}{\boldmath$0.9848$} & \textcolor{blue}{\boldmath$0.6558$} & \textcolor{blue}{\boldmath$0.9982$} & $0.1993$ \\

\vspace{0.15cm} & StyleGAN2 \cite{stylegan} & $0.4098$ & \textcolor{blue}{\boldmath$0.3846$} & $0.2577$ & $0.9821$ & $0.4603$ & $0.9926$ & $0.3694$ \\

\vspace{0.1cm} & Tile-StyleGAN2 \cite{stylegan} & \textcolor{red}{\boldmath$0.4373$} & \textcolor{red}{\boldmath$0.3980$} & \textcolor{red}{\boldmath$0.2799$} & $0.9838$ & $0.5234$ & $0.9942$ &	\textcolor{red}{\boldmath$0.3755$} \\ \bottomrule
\end{tabular}
}
\end{table*}

The best results are achieved by the proposed ADA-Net approaches when the Tile-StyleGAN2 discriminator configuration is used, resulting in $0.4373$ in dice score and $0.3980$ in $F_2-$score as presented in Table \ref{tab:results}. There is a significant performance gap between the best results achieved by the ADA-Net and the competing methods. For example, comparing the proposed approach and the best-obtained competing results, i.e., CUT with Tile-StyleGAN2 discriminator, the differences in the performances are approximately $4.5\%$, $7\%$, and $3\%$ in dice score, $F_2-$score, and IoU, respectively. Notably, all methods are able to reach similar performance levels in accuracy and specificity, whereas the best precision has been achieved by the FastCUT method. On the other hand, that particular configuration has obtained only $17\%$ sensitivity. In this application, our primary aim is to identify dead trees where it is preferred to have increased sensitivity levels with reasonable specificity and precision. Therefore, we mainly focus on dice scores and $F_2-$scores in our experimental evaluations. Finally, it is worth mentioning that while the Cycle-GAN approach cannot achieve better dice scores than the CUT model, its performance is comparably less affected by different discriminator configurations.

In the subsequent experimental evaluation, the segmentation performances are presented without applying domain adaptation, where the pre-trained segmentation network, Flair U-Net, has been directly inferred over the USA data. Accordingly, comparing the baseline results with Table \ref{tab:base_results}, all domain adaptation methods have increased performance levels in Table \ref{tab:results}. However, the proposed domain adaptation approach significantly increases the dice score by nearly $20$ percentage points and achieves an $F_2-$score that is more than two times compared to the baseline results obtained without domain adaptation. It is worth noting that the standing dead tree segmentation is a very challenging task considering the Flair U-Net segmentation model can only achieve less than $75\%$ in dice score over the Finland dataset, even though the network is originally trained using a large amount of annotated data from the Finland site. Therefore, another objective of this work, which was to discover the feasibility of using domain adaptation for zero-shot standing dead tree segmentation over different domain images, has been addressed by showing reasonable performance levels obtained by adapting source domain USA images into the Finland domain.

\begin{table}[]
\centering
\caption{Obtained dead tree segmentation results by Flair U-Net model over source and target domain images, i.e., USA and Finland, respectively.}
\label{tab:base_results}
\setlength{\tabcolsep}{18pt}
\renewcommand{\arraystretch}{1.4}
\resizebox{0.85\linewidth}{!}{
\begin{tabular}{@{}ccc@{}}
\toprule
            & \multicolumn{2}{c}{\textbf{Dataset}} \\ \cmidrule(l){2-3} 
            & \textbf{USA*}        & \textbf{Finland}      \\ \midrule
\textbf{Dice Score}  & $0.2436$       & $0.7380$       \\
\boldmath$F_2-$\textbf{Score}    & $0.1734$       & $0.7808$       \\
\textbf{IoU}         & $0.1387$       & $0.5848$       \\
\textbf{Accuracy}    & $0.9848$       & $0.9977$       \\
\textbf{Precision}   & $0.750$       & $0.6764$       \\
\textbf{Specificity} & $0.9992$       & $0.9984$       \\
\textbf{Sensitivity} & $0.1454$       & $0.8121$       \\ \bottomrule
\multicolumn{1}{c}{}         & \multicolumn{2}{r}{\scriptsize\textbf{*without domain adaptation}}
\end{tabular}
}
\end{table}

\begin{figure*}[h]
    \centering
    \includegraphics[width=1\linewidth]{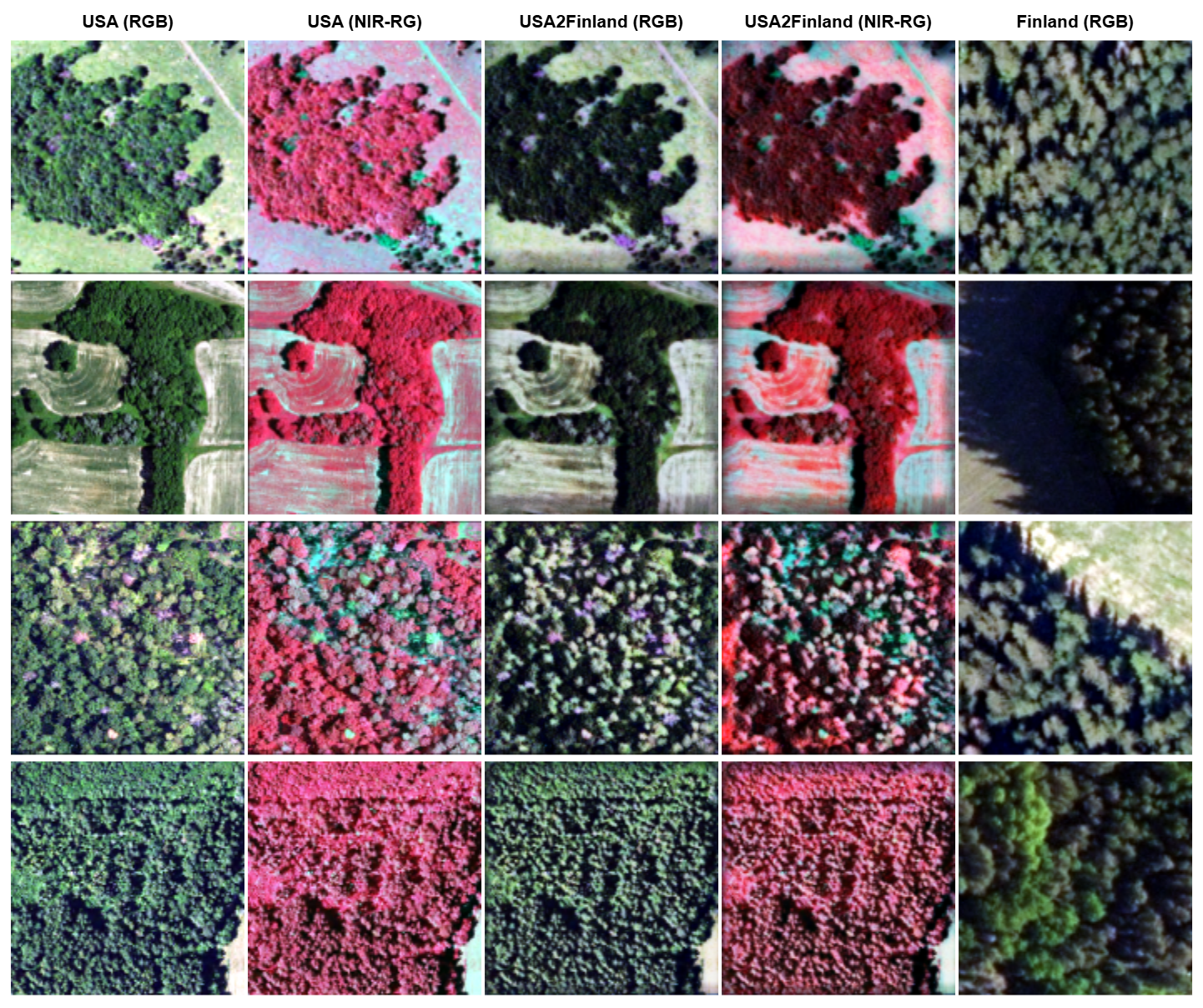}
    \caption{Four sample images are provided from the source domain (USA) and their transformations (USA2Finland) using the proposed ADA-Net approach. They are illustrated in red, green, blue (RGB) and false color illustration where near-infrared (NIR), red, and green channels are assigned to RGB, respectively. Randomly selected samples from the Finland domain are provided for comparison.}
\label{fig:transformation_results}
\end{figure*}

For visual evaluation of the segmented regions of the standing dead trees, we provide the Flair U-Net segmentation results over the original USA and USA2Finland images generated by the domain adaptation approaches in Fig. \ref{fig:segmentation_results}. It is demonstrated that when the pre-trained network is applied using the source domain images, many dead trees are missed in the segmentation. For example, in the first-row image of Fig. \ref{fig:segmentation_results}, without domain adaptation, no dead trees have been detected at all over the USA data. Generally, every approach has increased the performance to some degree, but the best segmentation results are obtained using the proposed domain adaptation technique. Considering the third given example image, the middle column of Fig. \ref{fig:segmentation_results}, it is observed that the ADA-Net approach is able to preserve certain characteristics: while Cycle-GAN generated the water region in blue color, the same area has a more dark appearance in the transformed image by the CUT method. On the other hand, the ADA-Net approach provides blackish/brown water appearance which is more realistic in Finland due to the organic rich natural waters with humus and because of high iron concentration.

Additional visual results are given in Fig. \ref{fig:transformation_results}, where the original source domain images, their domain-adapted versions, and random samples from the target domain images are illustrated. Original source domain images tend to be brighter in general compared to the images from Finland because of the illumination conditions of the two countries. Therefore, the domain-adapted USA2Finland images appear darker showing similar characteristics with Finland images, yielding overall better segmentation accuracies when a pre-trained network is trained over Finland images as previously discussed in the quantitative evaluations. For instance, images of trees in the Finland site generally have larger shadows and non-illuminated regions. This might be attributed to the fact that images are acquired by significantly different observation angles or geographical location constraints of being located in the northern region. Different solar incidence angles cause variations in image quality and characteristics between two locations. Observing the false color representations that are obtained by assigning NIR, R, G channels to R, G, B for the illustration, existing dead trees in the original images are preserved after the applied domain adaptation. In fact, thanks to the opposite contrast and darker overall appearance in the USA2Finland images, standing dead trees become more prominent and they are easier to identify through visual inspection than with the original domain images. It is worth mentioning that dead trees appear green in false color (NIR-RG) representation, while they are whitish in RGB.

\subsection{Computational Complexity Analysis}

Table \ref{tab:complexity} presents the computational complexities of methods in terms of number of trainable parameters and required elapsed times averaged per sample. Thanks to the followed compact configuration, the proposed approach has the smallest number of parameters across all discriminator configurations compared to other approaches. In contrast, the Cycle-GAN approach has the largest number of parameters due to the forward and inverse mapping tasks. On the other hand, only forward mapping generator is utilized during inference. As a result, all methods have obtained comparable elapsed times for inference including Cycle-GANs. Despite having fewer parameters, the ADA-Net approach has demonstrated similar inference elapsed times. This is caused by the involved attention mechanism which demands additional operations, hence equalizing the overall elapsed times with compared approaches.

\begin{table}[h!]
\centering
\caption{Number of total trainable parameters including generator and discriminator networks are given for each approach together with the elapsed times averaged per sample during inference (generator only).}
\label{tab:complexity}
\renewcommand{\arraystretch}{1.5}
\resizebox{\linewidth}{!}{
\begin{tabular}{|c|c|cc|}
\cline{1-4}
 \textbf{Method} & \multicolumn{1}{c|}{\textbf{Discriminator}} & \multicolumn{1}{c}{\textbf{Parameters}} & \multicolumn{1}{c|}{\textbf{Time (sec)}}  \\ \hline

\multicolumn{1}{|c|}{\multirow{4}{*}{CUT \cite{cut}}} & PatchGAN \cite{isola2017image} & $14.149$ M & \multicolumn{1}{|c|}{\multirow{4}{*}{0.2020}} \\ 

\multicolumn{1}{|c|}{} & PixelGAN & $11.393$ M & \multicolumn{1}{|c|}{} \\ 

\multicolumn{1}{|c|}{} & StyleGAN2 \cite{stylegan} & $25.647$ M & \multicolumn{1}{|c|}{} \\ 

\multicolumn{1}{|c|}{} & Tile-StyleGAN2 \cite{stylegan} & $25.647$ M & \multicolumn{1}{|c|}{} \\ \hline

\multicolumn{1}{|c|}{\multirow{4}{*}{FastCUT \cite{cut}}} & PatchGAN \cite{isola2017image} & $14.149$ M & \multicolumn{1}{|c|}{\multirow{4}{*}{0.2024}} \\ 

\multicolumn{1}{|c|}{} & PixelGAN & $11.393$ M & \multicolumn{1}{|c|}{} \\ 

\multicolumn{1}{|c|}{} & StyleGAN2 \cite{stylegan} & $25.647$ M & \multicolumn{1}{|c|}{} \\ 

\multicolumn{1}{|c|}{} & Tile-StyleGAN2 \cite{stylegan} & $25.647$ M & \multicolumn{1}{|c|}{} \\ \hline

\multicolumn{1}{|c|}{\multirow{4}{*}{Cycle-GAN \cite{cycle_gan}}} & PatchGAN \cite{isola2017image} & $28.298$ M & \multicolumn{1}{|c|}{\multirow{4}{*}{0.2394}} \\ 

\multicolumn{1}{|c|}{} & PixelGAN & $22.786$ M & \multicolumn{1}{|c|}{} \\ 

\multicolumn{1}{|c|}{} & StyleGAN2 \cite{stylegan} & $51.294$ M & \multicolumn{1}{|c|}{} \\ 

\multicolumn{1}{|c|}{} & Tile-StyleGAN2 \cite{stylegan} & $51.294$ M & \multicolumn{1}{|c|}{} \\ \hline 

\multicolumn{1}{|c|}{\multirow{4}{*}{\makecell{\textbf{ADA-Net} \\ \textbf{(Proposed)}}}} & PatchGAN \cite{isola2017image} & $9.724$ M & \multicolumn{1}{|c|}{\multirow{4}{*}{0.2085}} \\ 

\multicolumn{1}{|c|}{} & PixelGAN & $6.968$ M & \multicolumn{1}{|c|}{} \\ 

\multicolumn{1}{|c|}{} & StyleGAN2 \cite{stylegan} & $21.222$ M & \multicolumn{1}{|c|}{} \\ 

\multicolumn{1}{|c|}{} & Tile-StyleGAN2 \cite{stylegan} & $21.222$ M & \multicolumn{1}{|c|}{} \\ \hline

\end{tabular}}
\end{table}

\subsection{Limitations and Discussion}

Tree mortality detection is essential for environmental monitoring to preserve forest ecosystems. Simply, its ecological benefits are beyond the obvious: support for a sustainable forest management, contributing to carbon deposits, maintaining the habitats for wildlife, regulating water cycles, and overall resilient forests with proper tree mortality monitoring is directly linked with climate stability conservation. To this end, several methods have been proposed utilizing active remote sensing such as LiDAR data for tree mortality mapping. On the other hand, such data on large scales are practically not available due to its acquisition cost and accessibility issues. To address this drawback, standing dead tree segmentation is performed using aerial imagery in this work.

While recent advances in machine learning have enabled state-of-the-art performance levels in different computer vision applications, a tremendous amount of training data is required to train complex deep networks. This requirement limits the feasibility of using them in real-world applications. For example, one can argue that the segmentation performances achieved after the domain adaptation are not entirely satisfactory. On the other hand, the segmentation network has not been exposed to the USA data or the labels; consequently, such a segmentation approach is ``zero-shot" segmentation. In the scope of zero-shot segmentation, around $44\%$ dice score is sufficient in this application. Furthermore, even though the estimated canopies are not truly accurate in Fig. \ref{fig:segmentation_results} because of the segmentation, the primary aim is to detect standing dead trees. Thus, overlapped regions between the estimated and true tree masks are of limited importance; as long as only a few pixels are detected, we can consider the objective has been accomplished. When demonstrating the visual results, we have noticed that labels are not strictly accurate. For instance, the first example case in Fig. \ref{fig:segmentation_results} should have more annotated standing dead trees than what is shown in the true mask, meaning that some dead trees were missed by the forest expert annotators. Furthermore, many annotated polygons lack precise drawings affecting the overall segmentation performances reported in this study. As aforementioned, the detection of the standing dead trees from aerial images is a very challenging task that even forest experts were unable to identify straight without careful consideration. The task is specifically hard in the USA dataset for the segmentation network. This is because the annotation also includes brown trees that have died more recently. Browning is particularly visible in pine trees and they can remain brown for a few years since their needles are retained for some period after death. However, spruces in boreal forests have a brief browning phase unlike pines as their needles fall off shortly after becoming brown.

\section{Conclusion}
\label{sec:conclusion}

In this study, we have proposed utilizing multi-spectral aerial imagery for standing dead tree segmentation. Moreover, this study explores the practicality of using domain adaptation techniques where the transformation from source to target domain is learned through Generative Adversarial Networks (GANs). Specifically, the USA images are selected as source domain images, for which there is only a limited amount of annotated data, making it impractical to train the segmentation network. The target domain is selected as the Finland domain. A Flair U-Net model is trained with a sufficient volume of available labels for aerial images captured over Finland. Subsequently, the USA images are transformed to obtain \textit{USA2Finland} images which enables effective deployment of the pre-trained segmentation network. The explained Image-to-Image translation problem is challenging since the USA images differ significantly from the Finland images considering tree species, geographical differences, and different image acquisition angles. Overall, we show that satisfactory segmentation performance levels have been achieved for a zero-shot cross-site segmentation task.

Furthermore, we have proposed Attention-guided Domain Adaptation Network (ADA-Net) with an improved contrastive learning objective combining pixel-wise spatial and patch-wise frequency domain representation in the calculated contrastive losses. The proposed approach incorporates residual self-attention mechanisms in a compact configuration having substantially less trainable parameters than other approaches. Despite its compactness, it has obtained the best results outperforming the competing domain adaptation approaches. Finally, the annotated data for the USA images, labeled by our collaborating forest experts, have been publicly shared with the research community. This substantial contribution will naturally attract further research in tree mortality mapping.

In this work, the ADA-Net is used for cross-site evaluation with domain adaptation targeting standing dead tree segmentation. However, the proposed approach is a generic domain adaptation technique, indicating its potential for a broad range of applications. This will be the topic of our future work, where we plan to explore other domain adaptation challenges further and investigate the adaptability of the ADA-Net in these tasks.

\section*{Acknowledgments}

This work was funded by the European Union (ERC-2023-STG grant agreement no. 101116404). Views and opinions expressed are, however, those of the author(s) only and do not necessarily reflect those of the European Union or the European Research Council Executive Agency. Neither the European Union nor the granting authority can be held responsible for them. The authors greatly acknowledge CSC – IT Center for Science, Finland, for computational resources.

\ifCLASSOPTIONcaptionsoff
  \newpage
\fi



%

\bibliographystyle{IEEEtran}
\bibliography{IEEEtran}




\end{document}